\documentclass[runningheads]{llncs}
\usepackage{graphicx}
\usepackage{multirow}
\usepackage[table]{xcolor}
\usepackage{colortbl}
\usepackage{amsmath}
\usepackage{cuted}
\usepackage{amssymb}
\usepackage{float}
\usepackage{hyperref}
\usepackage{marvosym}

\begin{document}
\title{ITVTON: Virtual Try-On Diffusion Transformer 
 \\ Based on Integrated Image and Text}
\titlerunning{ITVTON}
%
\author{Haifeng Ni \inst{(}\textsuperscript{\raisebox{-0.18ex}{\scalebox{1.4}{\Letter}}}\inst{)} \and
Ming Xu}
\authorrunning{Haifeng Ni, Ming Xu}
%
\institute{School of Software Engineering, East China Normal University, Shanghai, China \\
\email{71255902129@stu.ecnu.edu.cn, mxu@cs.ecnu.edu.cn}}
%
%
\maketitle  
\begin{figure*}
    \centering
    \includegraphics[width=\textwidth]{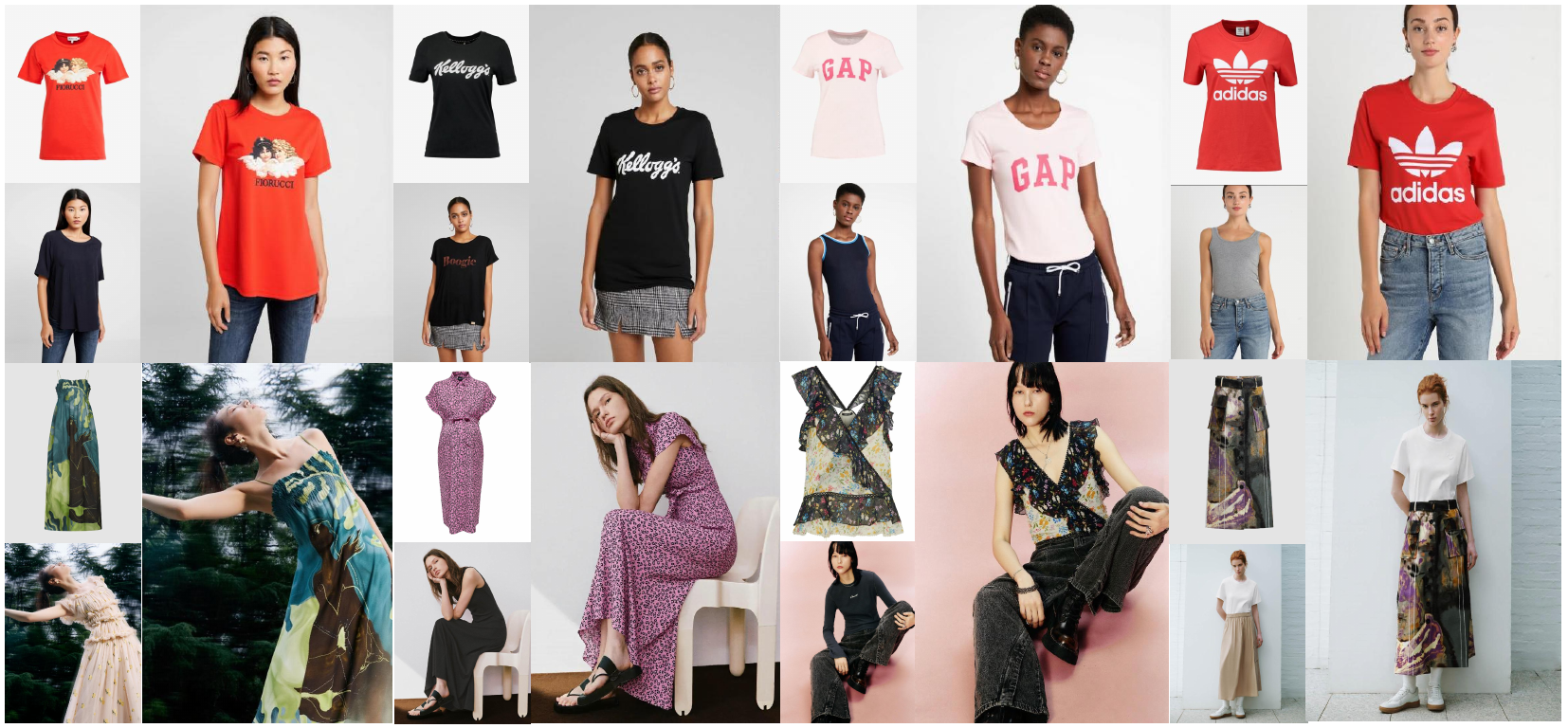}
    \caption{The ITVTON model is utilized to create virtual try-on images derived from the VITON-HD~\cite{Choi_2021_CVPR} dataset (row 1) and the filtered IGPair~\cite{shen2024imagdressingv1customizablevirtualdressing} dataset (row 2). For optimal visual evaluation, it is recommended to examine the images in enlarged form.}
\end{figure*}

\vspace{-12pt}
\begin{abstract}
Virtual try-on, which aims to seamlessly fit garments onto person images, has recently seen significant progress with diffusion-based models. However, existing methods commonly resort to duplicated backbones or additional image encoders to extract garment features, which increases computational overhead and network complexity. In this paper, we propose ITVTON, an efficient framework that leverages the Diffusion Transformer (DiT) as its single generator to improve image fidelity. By concatenating garment and person images along the width dimension and incorporating textual descriptions from both, ITVTON effectively captures garment-person interactions while preserving realism. To further reduce computational cost, we restrict training to the attention parameters within a single Diffusion Transformer (Single-DiT) block. Extensive experiments demonstrate that ITVTON surpasses baseline methods both qualitatively and quantitatively, setting a new standard for virtual try-on. Moreover, experiments on 10,257 image pairs from IGPair confirm its robustness in real-world scenarios.

\keywords{virtual try-on  \and diffusion transformer \and parameter training}
\end{abstract}

\section{Introduction}
Virtual try-on technology~\cite{Chen_2023_ICCV,Dong_2019_ICCV,gao2024exploring} has emerged as an indispensable tool in e-commerce and personal fashion, significantly enhancing consumers’ shopping experiences by generating realistic try-on images and offering novel marketing avenues for brands and retailers. Traditional methods often rely on Generative Adversarial Networks (GANs)~\cite{10.1145/3422622} in a two-stage process: first warping the garment according to the person’s pose, then using a generator to merge the warped garment with the person image. Although these approaches improve image realism, they heavily depend on the warping module, leading to potential overfitting, unstable generation quality, limited generalization, and difficulties in handling complex backgrounds or dynamic poses.

Recently, diffusion models~\cite{shen2024boosting,shen2025long} have emerged as a compelling alternative for virtual try-on tasks, owing to their superior generative capabilities and enhanced flexibility. Unlike GAN-based pipelines, diffusion models~\cite{shen2023advancing} tend to generate more stable, high-quality, and diverse outputs. They also possess strong control and optimization properties during training, making them well-suited for addressing challenges in virtual try-on. Current diffusion-based virtual try-on methods primarily adopt parallel U-shaped UNet structures~\cite{10.1007/978-3-319-24574-4_28}, typically introducing a garment UNet alongside a generative UNet, with attention mechanisms bridging garment features and the person image. While this approach substantially advances virtual try-on, it also entails additional network modules and higher computational costs at both training and inference stages.

Building upon the progress in diffusion transformers (DiT)~\cite{Peebles_2023_ICCV}, our goal is to harness their scalability and generative performance to improve virtual try-on image quality. To streamline the network design and jointly capture person and garment information, we avoid auxiliary image encoders by instead concatenating the garment and person images along the width dimension. As the DiT model capitalizes on textual cues, we further enhance realism by introducing textual descriptions of both the garment and person. Within the transformer backbone, image and text signals are fused directly through attention mechanisms, which proves critical for improving garment-person interactions. To reduce computational overhead, we focus training solely on the attention parameters within a single DiT block (Single-DiT), thereby striking a balance between performance and efficiency.

The contributions of the present paper are summarized as: 
\begin{itemize} \item[\textcolor{black}{$\bullet$}] We propose \emph{ITVTON}, a DiT-based virtual try-on model that leverages the strong scaling and generative capabilities of DiT to preserve fine-grained garment features and enhance interactions with person images. 
\item[\textcolor{black}{$\bullet$}] We align try-on results by concatenating garment and person images along the width dimension, integrating image-text descriptions for improved generation quality. This design eliminates the need for an extra image encoder, resulting in a compact network structure. \item[\textcolor{black}{$\bullet$}] We introduce a parameter-efficient training scheme that targets only the attention parameters within a Single-DiT block, reducing training time and computational overhead while enhancing stability and output fidelity. 
\end{itemize}

\section{Related Work}
The virtual try-on task~\cite{gao2024exploring} involves generating a realistic photograph of a model wearing a specified outfit while maintaining consistency in other regions. Scholars initially explored generative adversarial networks (GANs)\cite{10.1145/3422622} for this task, employing a two-stage process: aligning the garment with the model and fusing them via a GAN-based network. VITON-HD\cite{Choi_2021_CVPR} introduces ALIgnment-Aware Segment (ALIAS) normalization to address misalignment by segmenting and fitting garments. GP-VTON~\cite{xie2023gpvtongeneralpurposevirtual} proposes a Local-Flow Global-Parsing (LFGP) warping module and a dynamic gradient truncation strategy to simulate garment deformation more effectively. However, GAN-based models suffer from unstable training and struggle to preserve garment texture details.

Recently, diffusion models have demonstrated superior stability and quality in virtual try-on. LADI-VTON~\cite{10.1145/3581783.3612137} and DCI-VTON~\cite{10.1145/3581783.3612255} align garments with the body before using diffusion models for fusion. TryOnDiffusion~\cite{Zhu_2023_CVPR} employs a parallel UNet to unify garment deformation and fusion. OOTDiffusion~\cite{xu2024ootdiffusionoutfittingfusionbased} captures detailed clothing features in a single step, while IDM-VTON~\cite{choi2024improvingdiffusionmodelsauthentic} incorporates the IP-Adapter and pose guidance for enhanced control. Paint-by-Example~\cite{Yang_2023_CVPR} reframes try-on as an inpainting problem, fine-tuning diffusion models on virtual try-on datasets to enable fine-grained image control.

Much work has been inspired by the PCDMs~\cite{shen2023advancing} second-stage inpainting conditional diffusion model, which uses the concatenation of source conditions and target images and is gradually gaining popularity in virtual try-on tasks. For example, CatVTON~\cite{chong2024catvtonconcatenationneedvirtual}  similarly fine-tunes the restoration diffusion model but eliminates the text encoder and cross-attention block, training only the self-attention block to create a lightweight network structure. IC-LoRA~\cite{huang2024incontextloradiffusiontransformers} posits that the text-to-image diffusion transformer model~\cite{Peebles_2023_ICCV} is inherently contextually generative, enabling high-quality image generation through integrated captioning of multiple images using small datasets (e.g., 20–100 samples) with task-specific LoRA~\cite{hu2021loralowrankadaptationlarge} tuning. In this paper, we adopt the DiT model as a prior and achieve a high-quality, high-fidelity virtual try-on task through a simple network architecture and an efficient training strategy, which involves splicing garment and person images along the width dimension and using integrated image-text descriptions as inputs. Additionally, our approach further improves generation performance and training efficiency while maintaining a lightweight network structure.

\section{Proposed Method}
\subsection{Background on Diffusion Model}
Stable Diffusion. The stable diffusion~\cite{Rombach_2022_CVPR} is a prominent example of the latent diffusion model. It comprises a variational autoencoder (VAE) $\mathcal {E}_{\theta}$, a CLIP~\cite{pmlr-v139-radford21a} text encoder $\tau_{\theta}$, and a denoising UNet $\epsilon_{\theta}$. The key advantage is that image information is compressed into a low-dimensional latent space, which significantly reduces the computational cost associated with training and inference. The input consists of an image $\mathbf {x}$ along with textual cues $\mathbf {y}$, and the following loss function is minimized during training:
\begin{equation}
\mathcal {L}_{LDM} = \mathbb {E}_{\mathcal {E}_{\theta}(\mathbf {x}),\mathbf {y},\epsilon \sim \mathcal {N}(0, 1),\mathbf {t}}\left [\lVert \epsilon - \epsilon _{\theta }(\mathbf {Z}_\mathbf {t}, \mathbf {t}, \tau_{\theta}(\mathbf {y}))\rVert _2^2\right ], 
\end{equation}
where $\mathbf {t} \in \{1, \ldots, T\}$ represents the time step of forward noise addition in the diffusion model,
and $\mathbf{Z}_\mathbf{t}$ denotes the VAE-encoded image with added random noise 
$\epsilon \sim \mathcal {N}(0, 1)$.

FLUX. FLUX is a type of diffusion model that employs the DiT architecture and is trained using a parameterize rectified flow model (RF)~\cite{esser2024scaling}. The training process of FLUX is defined as a rectified flow that links the data distribution to the noise distribution via a straight line. A key characteristic of the forward process is that $\mathbf {z}_\mathbf {t}$ is obtained by linearly interpolating the data $\mathbf {x}_0$ and the noise $\epsilon$, i.e.:
\begin{equation}
\mathbf {z}_\mathbf {t} = (1 - \mathbf {t})\mathbf {x}_0 + \mathbf {t}\epsilon,\epsilon \sim \mathcal {N}(0, 1), 
\end{equation}
The core concept of the rectified flow technique employed in FLUX is to streamline the training and inference processes by introducing a flow-matching method that enhances the efficiency of image generation. The loss function is defined as
\begin{equation}
\mathcal {L}_{RF} = -\frac{1}{2}\mathbb {E}_{\mathbf {t} \sim \mathcal {U}(\mathbf {t}),\epsilon \sim \mathcal {N}(0, 1)}\left [\mathbf {w}_\mathbf{t}\lambda_\mathbf{t} \lVert \epsilon _{\Theta}(\mathbf {z}_\mathbf {t}, \mathbf {t})-\epsilon\rVert ^2\right ], 
\end{equation}
where $\lambda_\mathbf{t}$ corresponds to a signal-to-noise ratio and $\mathbf {w}_\mathbf{t}$ is a time dependent-weighting factor. $\epsilon _{\Theta}(\mathbf {z}_\mathbf {t}, \mathbf {t})$ denotes the network parameters of the transformer module.

\begin{figure*}[t]
    \centering
    \includegraphics[width=\textwidth]{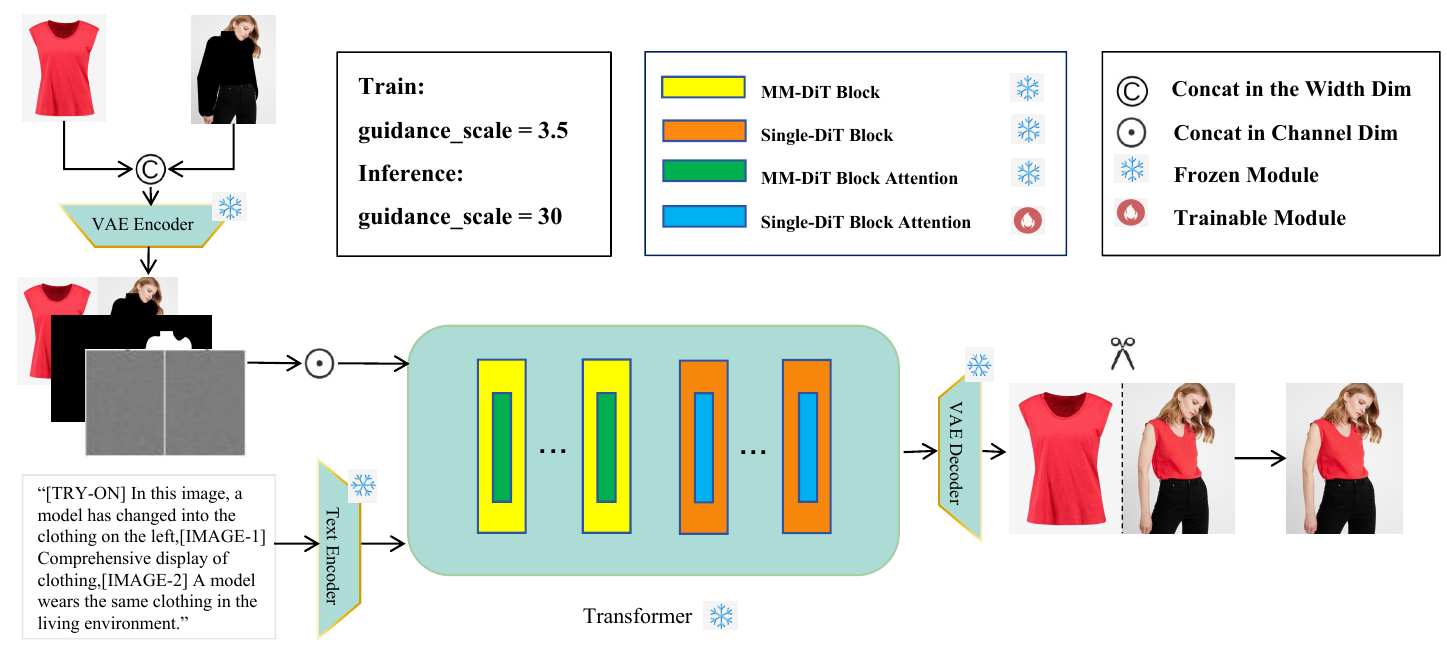}
    \caption{Overview of ITVTON. Our approach achieves high-quality virtual try-on by concatenating the garment image with the target person image along the width dimension and incorporating integrated image-text representations(as illustrated in the figure). Only the attention parameters in the Single-DiT module remain learnable during training, ensuring a streamlined and efficient try-on network.}
    \label{fig:framework}
\end{figure*}
\subsection{ITVTON}
The ITVTON architecture, illustrated in Fig.~\ref{fig:framework}, aims to generate high-quality and high-fidelity virtual try-on images using a simple network structure and minimal training cost. This section outlines the components of our network architecture and the underlying architectural rationale, as well as explores efficient parameter training strategies.

\subsubsection{Network Module} Previous approaches primarily focus on achieving detailed alignment between garments and persons. For instance, LADI-VTON~\cite{10.1145/3581783.3612137} first warps the garments to match the model's pose, then uses a diffusion model to blend the warped garments with the model. IDM-VTON~\cite{choi2024improvingdiffusionmodelsauthentic} employs a parallel UNet to extract garment features and incorporates an attention mechanism into the generative UNet to maintain alignment during virtual try-on. However, these methods either lack sufficient accuracy or necessitate the introduction of additional network modules for their implementation, failing to meet our expectations.

We found that the mature generative model exhibits a strong ability to generate masked regions; for example, when the mouth of a face is masked, the model is likely to generate an intact face upon repair. Based on this, if the clothing area of the model is masked and subsequently fed into the model along with a garment image and a mask image, what would happen during repair? The model would likely generate content for the masked area based on the image information outside the mask. However, to generate a high-quality and high-fidelity image of the garment change, the model must also be trained and fine-tuned. Furthermore, because the attention mechanism of FLUX’s transformer module processes image information in conjunction with textual data, it is beneficial to leverage the model’s textual input. 

Consequently, we directly incorporate the three modules:

\textbf{VAE}: The VAE encoder transforms pixel-level images into a latent representation to optimize computational efficiency. Compared to traditional VAEs, the FLUX series VAE outputs latent features. Before these features are fed into the diffusion model, a patching operation is performed, stacking $2\times2$ pixel blocks directly along the channel dimension. This method preserves the original resolution of each pixel block and prevents the loss of critical image feature information. Consequently, we concatenate the garment and person features along the width dimension, leveraging the robust capabilities of the VAE while ensuring alignment between inputs.

\textbf{Text Encoder}: Two text encoders (CLIP ViT-L and T5-XXL~\cite{JMLR:v21:20-074}) process textual prompts as input, enabling the generation of more controllable, high-quality images. Inspired by IC-LoRA~\cite{huang2024incontextloradiffusiontransformers}, multi-image integrated text inputs have a positive impact on virtual try-on tasks. Therefore, we utilize multi-image integrated text (text format shown in Fig.~\ref{fig:framework}) as input, enhancing the controllability and quality of the generated images.

\textbf{Transformer}: The transformer synthesizes the final try-on image by integrating features from the latent space. It accepts concatenated garment and person features, along with noise and masks as image inputs, and receives multi-image integrated titles as text inputs. By seamlessly integrating all this information, the transformer facilitates effective learning and synthesis of the final try-on image.

Our network architecture does not rely on additional modules, and the try-on task can be efficiently performed using only the inherent functionality of the model.

\subsubsection{Architectural Inference} In this work, we utilize stitched garment and person images as input and employ integrated image-text pairs as the textual input. This approach ensures a streamlined network architecture and simplifies the preprocessing steps prior to inference.

Specifically, consider a garment image $\mathbf {I}_\mathbf {g} \in \mathbb{R}^{3\times H \times W}$ and a person image $\mathbf {I}_\mathbf {p} \in \mathbb{R}^{3\times H \times W}$ , which are concatenated along the width dimension to generate $\mathbf {C}_\mathbf {gp} \in \mathbb{R}^{3\times H \times 2W}$ . The corresponding binary, clothing-agnostic mask map of the person image, denoted $\mathbf {m}_\mathbf {p} \in \mathbb{R}^{H \times W}$, is concatenated with an all-zero mask map of the same dimensions to produce $\mathbf {m}_\mathbf {op} \in \mathbb{R}^{H \times 2W}$.
\begin{gather}
\mathbf {C}_\mathbf {gp} = \mathbf {I}_\mathbf {g} \text{\copyright}  \mathbf {I}_\mathbf {p}, \\
\mathbf {m}_\mathbf {op} = \mathbf {O} \text{\copyright} \mathbf {m}_\mathbf {p}, 
\end{gather}
where $\text{\copyright}$ denotes concatenation along the width dimension, $\mathbf {O}$ represents an all-zero mask image.
$\mathbf {C}_\mathbf {gp}$ is multiplied with the processed elements of $\mathbf {m}_\mathbf {op}$ to generate the clothing-agnostic person and garment-spliced image $\mathbf {C}_\mathbf {masked} \in \mathbb{R}^{3\times H \times 2W}$. $\mathbf {C}_\mathbf {masked}$ is then encoded using the VAE encoder $\mathcal {E}$ to obtain $\mathbf {V}_\mathbf {masked} \in \mathbb{R}^{16\times \frac{H}{8} \times \frac{2W}{8}}$, which has a channel value of 16. Finally, $\mathbf {m}_\mathbf {op}$ is interpolated to produce $\mathbf {X}_\mathbf {om} \in \mathbb{R}^{64\times \frac{H}{8} \times \frac{2W}{8}}$ with a channel value of 64.
\begin{gather}
\mathbf {C}_\mathbf {masked} = \mathbf {C}_\mathbf {gp} \otimes  (1-\mathbf {m}_\mathbf {op}), \\
\mathbf {V}_\mathbf {masked}  = \mathcal {E}(\mathbf {C}_\mathbf {masked}), 
\end{gather}
where $\otimes$ denotes element-wise multiplication.$\mathbf {V}_\mathbf {masked}$ and 
$\mathbf {X}_\mathbf {om}$ undergo a packing operation to generate $\mathbf {P}_\mathbf {masked}  \in \mathbb{R}^{64\times (\frac{H}{8\times2} \times \frac{2W}{8\times2})}$ and $\mathbf {P}_\mathbf {om} \in \mathbb{R}^{256\times (\frac{H}{8\times2} \times \frac{2W}{8\times2})}$, respectively, before being fed into the transformer module.The channel values are scaled by a factor of 4, while the width and height are each halved, then multiplied and compressed into a single-dimensional value.
\begin{gather}
\mathbf {P}_\mathbf {masked} = pack(\mathbf {V}_\mathbf {masked}), \\
\mathbf {P}_\mathbf {om}  = pack(\mathbf {X}_\mathbf {om}), 
\end{gather}
The text encoder $\tau$ processes the integrated image-text pair, producing 
$\tau_{text}$. Prior to noise reduction, $\mathbf {P}_\mathbf {masked}$ and $\mathbf {P}_\mathbf {om}$ are concatenated along the channel dimension with random noise $\mathbf {Z}_\mathbf {t}$ of the same size as $\mathbf {P}_\mathbf {masked}$, which is then fed into the transformer module to predict $\mathbf {Z}_\mathbf {t-1}$ alongside $\tau_{text}$.
\begin{equation}
\mathbf {Z}_\mathbf {t-1} = Transformer(\mathbf {Z}_\mathbf {t} \odot \mathbf {P}_\mathbf {masked} \odot \mathbf {P}_\mathbf {om},\tau_{text}),  
\end{equation}
where $\odot$ denotes the splicing operation along the width dimension. $\mathbf {Z}_\mathbf {0} \in \mathbb{R}^{64\times (\frac{H}{8\times2} \times \frac{2W}{8\times2})}$ is obtained after t-steps of cyclic noise reduction, followed by 
$\mathbf {V}_\mathbf {con} \in \mathbb{R}^{16\times \frac{H}{8} \times \frac{2W}{8}}$, which is generated after the unpacking operation.
Next, $\mathbf {I}_\mathbf {con} \in \mathbb{R}^{3\times H \times 2W}$ is derived through VAE decoding.
The final image $\mathbf {I}_\mathbf {result} \in \mathbb{R}^{3\times H \times W}$ is produced after cropping.

\subsubsection{Parameter-Efficient Training} The pre-trained DiT model demonstrates strong capability and robustness in generating masked regions; however, fine-tuning is required to ensure accurate interaction between garment and person features, thereby generating efficient images. On the other hand, these pre-trained models already possess a wealth of prior knowledge, and an excessive number of trainable parameters not only increases the training burden—requiring more GPU memory and a longer training duration—but may also degrade the model’s existing performance.

The Transformer in FLUX comprises the MM-DiT block structures and the Single-DiT block structures. The MM-DiT block facilitates the fusion of the two modal information streams, which are subsequently fed into the Single-DiT block to deepen the model's architecture and enhance its learning capability. Additionally, a parallel attention mechanism is incorporated into the Single-DiT blocks to further optimize model performance. The attention within these blocks fuses both image and text information for processing. We hypothesized that the attention mechanism is crucial for achieving high-quality results and conducted experiments to identify the most relevant blocks. Specifically, we configured the trainable components as the all attention layer, the attention layer in the MM-DiT block, and the attention layer in the Single-DiT block. The experimental results indicate that training all attention layers, the attention layer in the MM-DiT block, and the attention layer in the Single-DiT block can generate satisfactory trial-wear effects; however, the metrics obtained by training the attention layer in the Single-DiT block are optimal, with the lowest number of training parameters. Therefore, we employ a parameter-efficient training strategy to fine-tune the attention layer within the Single-DiT block, which contains 1076.2M parameters. Additionally, we apply a 10\% probability to discard multi-image integrated text cues during training to improve the model's generalization ability.

\section{Experimentation and Analysis}

\subsection{Datasets}
Our experiments utilized two publicly available fashion datasets: VITON-HD~\cite{Choi_2021_CVPR} and IGPair~\cite{shen2024imagdressingv1customizablevirtualdressing}. The VITON-HD~\cite{Choi_2021_CVPR} dataset comprises 13,679 image pairs, including 11,647 pairs in the training set and 2,032 pairs in the test set. Each pair consists of a frontal upper-body image of a person paired with an upper-body clothing image, both at a resolution of $768\times1024$. The IGPair~\cite{shen2024imagdressingv1customizablevirtualdressing} dataset contains 324,857 image pairs with clothing classified into 18 categories encompassing a broad spectrum of styles, such as casual, formal, sports, fashion, and vintage. For each apparel item, 2 to 5 model images taken from various angles are provided, with a resolution of $1024\times1280$. From these, we meticulously selected 10,381 image pairs featuring the same model in multiple poses while wearing the same attire. Of these pairs, 10,257 are allocated for training purposes, with the remainder reserved for testing. We employ OpenPose~\cite{Cao_2017_CVPR,Simon_2017_CVPR,Wei_2016_CVPR} and HumanParsing~\cite{9310358} to generate garment-independent masks and uniformly adjust the resolution of all images to $768\times1024$ pixels.

\subsection{Implementation Details}
We utilize the pre-trained FLUX.1-Fill-dev model for training. We trained the model on the VITON-HD~\cite{Choi_2021_CVPR} dataset and evaluated its performance against other approaches using a designated test set. Additionally, we trained the model on the IGPair~\cite{shen2024imagdressingv1customizablevirtualdressing} dataset to assess its generative capability in realistic scenarios. Each model was trained with identical hyperparameters: 36,000 steps at a resolution of $384\times512$ using the AdamW 8-bit optimizer, a batch size of 4, and a constant learning rate of 1e-5. All experiments were conducted on a single NVIDIA A100 GPU.

\begin{figure*}[t]
    \centering
    \includegraphics[width=0.8\linewidth]{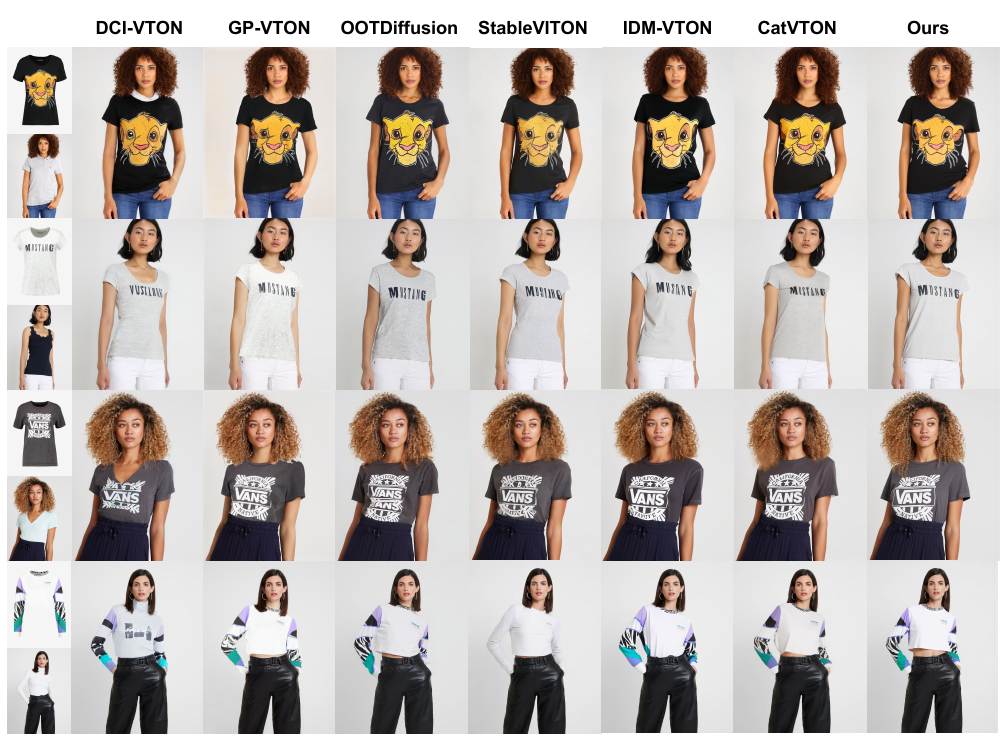}
    \caption{Qualitative comparison on the VITON-HD~\cite{Choi_2021_CVPR} Dataset. ITVTON exhibits significant advantages in processing complex patterns and text. We recommend zooming in for a detailed inspection.}
    \label{fig:comparison_chart}
\end{figure*}
\subsection{Qualitative Comparison}

We evaluated the visual quality using the VITON-HD~\cite{Choi_2021_CVPR} dataset. As shown in Fig.~\ref{fig:comparison_chart}, it demonstrates the try-on effects of various styles and patterns from the dataset, and compares the try-on variations across different methods. While other methods encounter issues such as color discrepancies and inconsistent details during virtual try-ons, ITVTON delivers more realistic and detailed outcomes owing to its simplicity and efficiency.

Additionally, we assess ITVTON on selected IGPair~\cite{shen2024imagdressingv1customizablevirtualdressing} datasets and benchmark it against other diffusion-based VTON methods. As shown in Fig.~\ref{fig:comparison_in_the_wild}, ITVTON demonstrates its capability to accurately recognize and model the shape of complex garments, such as a sheath dress, and seamlessly integrate them with human figures. Notably, it can generate high-quality images even in complex poses (e.g., sitting on the chair). In intricate field scenes, ITVTON effectively handles the integration of backgrounds with clothing, maintaining consistency and realism across diverse and challenging environments.

\subsection{Quantitative Comparison}

For the inference results of the test pair dataset, we use two metrics, frechet inception distance (FID)~\cite{heusel2017gans} and kernel inception distance (KID)~\cite{bińkowski2021demystifyingmmdgans}, to compare the gap between real and generated images, and two metrics, structural similarity index (SSIM)~\cite{1284395} and learned perceptual image patch similarity (LPIPS)~\cite{Zhang_2018_CVPR} which are two metrics to measure the overall structural and perceptual similarity between images.

We conducted a quantitative evaluation of several state-of-the-art open-source virtual try-on methods using the VITON-HD~\cite{Choi_2021_CVPR} dataset. Comparisons were made in paired settings to assess the similarity between the synthesized results and ground truth, as well as the model's generalization performance. The results, presented in Table~\ref{tab:table1}, show that our method outperforms all others across all metrics. This demonstrates the efficacy of our simplified network architecture and parameter fine-tuning in virtual try-on tasks. Additionally, the data indicates that both CatVTON~\cite{chong2024catvtonconcatenationneedvirtual} and IDM-VTON~\cite{choi2024improvingdiffusionmodelsauthentic} demonstrated strong performance in virtual try-on tasks.

\subsection{Ablation Studies}
We conducted an ablation study to investigate the effects of several factors:
\begin{enumerate}
\item different trainable modules,

\begin{table}[t]
    \caption{Quantitative Comparison with Other Methods. We compare metric performance in paired settings on the VITON-HD~\cite{Choi_2021_CVPR} dataset. The best and second-best results are bolded and underlined, respectively.
    }
    \label{tab:table1}
    \centering
    \renewcommand{\arraystretch}{1.3}
    \resizebox{0.55\textwidth}{!}{
    \begin{tabular}{c|cccc}
    \hline
    Methods      & $SSIM \uparrow$            & $FID \downarrow$           & $KID \downarrow$            & $LPIPS \downarrow$           \\ \hline

    DCI-VTON~\cite{10.1145/3581783.3612255} & 0.8620          & 9.408          & 4.547          & 0.0606          \\
 StableVITON~\cite{Kim_2024_CVPR} & 0.8543          & 6.439          & 0.942          & 0.0905          \\
 GP-VTON~\cite{xie2023gpvtongeneralpurposevirtual} & 0.8701          & 8.726          & 3.944          & 0.0585          \\
    
    OOTDiffusion~\cite{xu2024ootdiffusionoutfittingfusionbased} & 0.8187          & 9.305          & 4.086          & 0.0876          \\
    IDM-VTON~\cite{choi2024improvingdiffusionmodelsauthentic}     & 0.8499          & 5.762          & 0.732          & 0.0603          \\
    CatVTON~\cite{chong2024catvtonconcatenationneedvirtual}      & \underline{0.8704}    & \underline{5.425}    & \underline{0.411}    & \underline{0.0565}    \\
    \rowcolor[HTML]{00D2CB} ITVTON (Ours) & \textbf{0.8734} & \textbf{5.064} & \textbf{0.264} & \textbf{0.0530} \\ \hline
    \end{tabular}
    }
    
    \end{table}
\begin{table}[t]
 \caption{Ablation results of different trainable module on VITON-HD~\cite{Choi_2021_CVPR} dataset.Taken together, the results indicate that training the attention parameters within the Single-DiT block achieves the optimal performance.The best results are shown in bold.
        }
        \label{tab:table2}
        \centering
        \renewcommand{\arraystretch}{1.3}
        \resizebox{0.9\textwidth}{!}{
        \begin{tabular}{c|cccc|c}
        \hline
        Trainable Module           & $SSIM \uparrow$            & $FID \downarrow$            & $KID \downarrow$            & $LPIPS \downarrow$           & Trainable Params(M) \\ \hline
        All Attention              & 0.8702    & 5.155    & \textbf{0.250} & 0.0557    & 1611.23             \\
        MM-DiT Block Attention     & 0.8641          & 5.438          & 0.332          & 0.0578          & 1434.93             \\
        \rowcolor[HTML]{00D2CB} Single-DiT Block Attention & \textbf{0.8734} & \textbf{5.064} & 0.264    & \textbf{0.0530} & 1076.2              \\ \hline
        \end{tabular}
        }
        
        \end{table}

\item the inclusion or exclusion of multi-image integrated text during both training and inference,
\item the impact of the guidance scale on the try-on results.
\end{enumerate}

For a fair comparison, we trained different versions of the model using the VITON-HD~\cite{Choi_2021_CVPR} dataset.

\textbf{Trainable Modules}: We evaluated three trainable modules in Transformer: \textup{1)} all attention parameters. \textup{2)} attention parameters of MM-DiT blocks. \textup{3)} attention parameters of Single-DiT blocks, with progressively fewer trainable parameters. Table~\ref{tab:table2} presents the evaluation results on the VITON-HD~\cite{Choi_2021_CVPR} dataset. We observe that the model trained with attention parameters within the Single-DiT block performs optimally across three metrics. The other metric, KID, shows only a slight decrease compared to the model that trains all attention parameters, while the Single-DiT model uses only 1076.2M attention parameters. Fewer training weights both reduce memory requirements and speed up training, and do not pull down performance. This demonstrates that training the attention parameters within the Single-DiT block is optimal.

\textbf{Multi-Image Integrated Text}: Ordinary text, such as ``A model is wearing a top.''. Multi-image integrated text is shown in Fig.~\ref{fig:framework}. Both types of text inputs were employed during training and inference. As shown in Table~\ref{tab:table3}, two metrics (SSIM and LPIPS) perform optimally when only multi-image integrated text is used during inference, although the improvement is marginal compared to the model using multi-image integrated text for both training and inference. Additionally, the other two metrics (FID and KID) are optimal for the model that utilizes multi-image integrated text for both training and inference, demonstrating that this approach is the most effective.

\begin{figure}[H]
  \centering
  \begin{minipage}[b]{0.5\textwidth}
    \includegraphics[width=\textwidth]{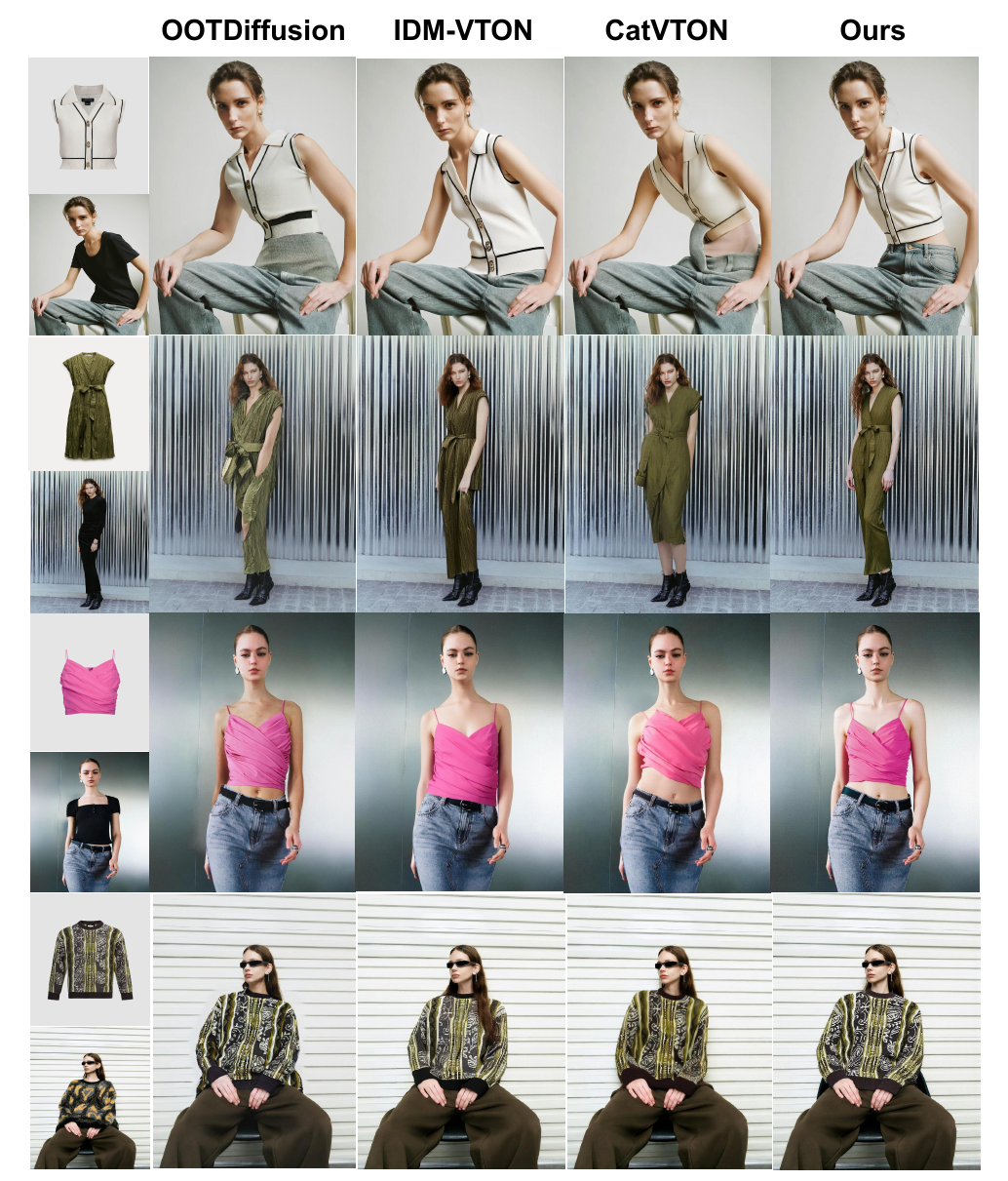}
    \caption{Qualitative comparison in field scenarios demonstrates that our method generates more natural try-on effects, even in complex scenes and with varied postures.}
    \label{fig:comparison_in_the_wild}
  \end{minipage}
  \hfill
  \begin{minipage}[b]{0.45\textwidth}
    \includegraphics[width=\textwidth]{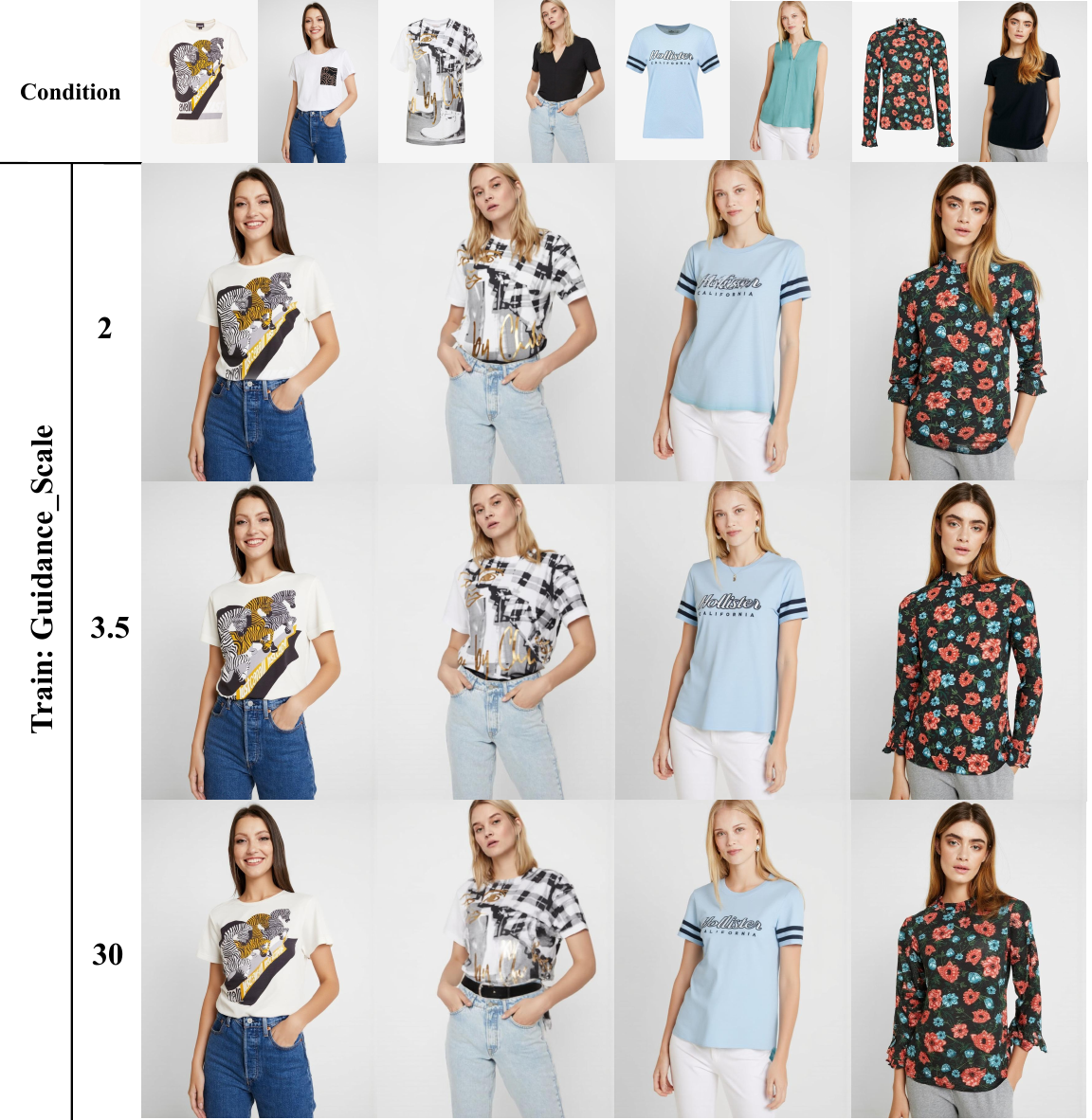}
    \caption{Comparison of the effects of setting different guidance scales during training. The model with the guidance scale set to 30 during inference and the guidance scale set to 2 or 30 during training does not perform well when generating garments with text or complex patterns. 
    }
    \label{fig:train_guidance_scale}
  \end{minipage}
\end{figure}

\begin{figure}[H]
     \centering
     \includegraphics[width=0.7\linewidth]{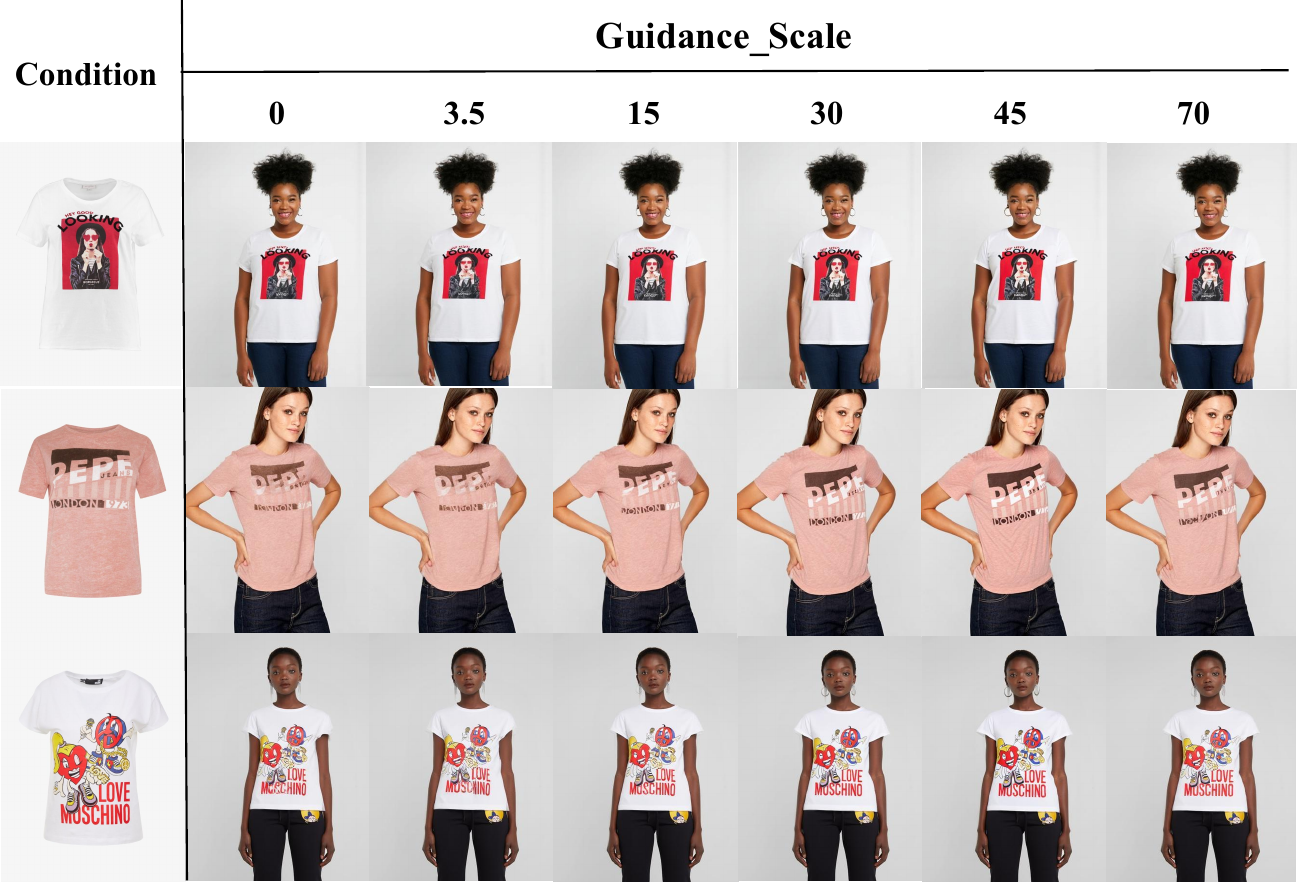}
      \caption{Comparison of the effects of setting different guidance scales during inference. 
      }
    \label{fig:inference_guidance_scale}
\end{figure}

\textbf{Guidance Scale}: The guidance scale is crucial in both the training and inference processes of the FLUX model. First, adjust the guidance scale setting to 30 for inference, as shown in Table~\ref{tab:table4}. The metrics FID, KID, and LPIPS achieve optimal results with a guidance scale setting of 3.5 during training, while the SSIM metric is slightly lower compared to the model with a guidance scale of 30 during training. This suggests that setting the guidance scale to 3.5 during training yields better performance. As shown in Fig.~\ref{fig:train_guidance_scale}, models with the guidance scale set to 2 or 30 during training do not perform well when generating garments with repeated or intricate patterns, while models with the guidance scale set to 3.5 can accurately reproduce repeated patterns or images. Next, set the guidance scale to 3.5 for training and evaluate the effects of different guidance scale settings during inference. As shown in Fig.~\ref{fig:inference_guidance_scale}, setting the guidance scale to 30 or 45 during inference results in the best restoration of color and pattern. For example, setting the guidance scale to 0 during inference results in color discrepancies and broken patterns. In summary, setting the guidance scale to 3.5 for training and 30 for inference yields the best inference results.



\begin{table}[t]
\caption{Ablation results from post-training inference on the VITON-HD~\cite{Choi_2021_CVPR} dataset using different text inputs are presented. The best results are shown in bold. Overall, training with integrated text and reasoning with integrated text yield the best performance.
            }
            \label{tab:table3}
\centering
\renewcommand{\arraystretch}{1.3}
\resizebox{0.9\textwidth}{!}{
\begin{tabular}{cc|cc|cccc}
\hline
\multicolumn{2}{c|}{Train}      & \multicolumn{2}{c|}{Inference}  & \multicolumn{1}{c|}{}                       & \multicolumn{1}{c|}{}                      & \multicolumn{1}{c|}{}                      &                         \\ \cline{1-4}
Ordinary Text & Integrated Text & Ordinary Text & Integrated Text & \multicolumn{1}{c|}{\multirow{-2}{*}{$SSIM \uparrow$}} & \multicolumn{1}{c|}{\multirow{-2}{*}{$FID \downarrow$}} & \multicolumn{1}{c|}{\multirow{-2}{*}{$KID \downarrow$ }} & \multirow{-2}{*}{$LPIPS \downarrow$} \\ \hline
\checkmark             & -               & \checkmark              & -               & 0.8727                                      & 5.178                                      & 0.377                                      & 0.0534                  \\
\checkmark              & -               & -             & \checkmark                & \textbf{0.8737}                             & 5.169                                      & 0.379                                      & \textbf{0.0529}         \\
-             & \checkmark                & \checkmark              & -               & 0.8723                                      &  5.098                               &  0.298                                & 0.0537                  \\
\rowcolor[HTML]{00D2CB} 
-             & \checkmark                & -             & \checkmark                &  0.8734                                & \textbf{5.064}                             & \textbf{0.264}                             &  0.0530           \\ \hline
\end{tabular}
}

\end{table}

\begin{table}[t]
\caption{Ablation results for various guidance scale settings during training. The best results are shown in bold. Setting the guidance scale to 30 during inference and to 2, 3.5, and 30 during training results in very similar values for the four metrics of the VITON-HD~\cite{Choi_2021_CVPR} test set. However, collectively, a guidance scale of 3.5 emerges as the optimal setting.
            }
            \label{tab:table4}
\centering
\renewcommand{\arraystretch}{1.3}
\resizebox{0.6\textwidth}{!}{
\begin{tabular}{c|cccc}
\hline
Guidance\_Scale & $SSIM \uparrow$           & $FID \downarrow$           & $KID \downarrow$            & $LPIPS \downarrow$           \\ \hline
2               & 0.8727          & 5.246          & 0.394         & 0.0532          \\
\rowcolor[HTML]{00D2CB}3.5             &  0.8734    & \textbf{5.064} & \textbf{0.264} & \textbf{0.0530} \\
30              & \textbf{0.8748} &  5.068    &  0.267    &  0.0540    \\ \hline
\end{tabular}
}

\end{table}

\section{Conclusion}
In this paper, we introduced ITVTON, a simple yet effective virtual try-on diffusion transformer model with 1,076.2M trainable parameters.
ITVTON spliced garment and person images along the width dimension, taken integrated image-text descriptions as input, and required only fine-tuning to generate high-quality, high-fidelity virtual try-on images.

Extensive experiments have demonstrated that ITVTON could achieve superior qualitative and quantitative results compared to state-of-the-art methods, while maintaining a simple and efficient network architecture. These findings also suggested that ITVTON would hold significant potential for widespread application in virtual try-on technology.

%
%
%
%
\bibliographystyle{splncs04}
\bibliography{arxiv}

\end{document}